%% file: arXiv2.tex
\documentclass[12pt]{article}

\usepackage[margin=1in]{geometry}

\usepackage{setspace}

\usepackage{graphicx}

\usepackage{enumitem}

\usepackage{microtype}
\usepackage{subfigure}
\usepackage{booktabs}
\usepackage{comment}
\usepackage{wrapfig}
\usepackage{arydshln}
\usepackage{enumitem}
\usepackage{multirow}
\usepackage{url}
\usepackage{natbib}
\usepackage{authblk}

\RequirePackage[colorlinks,citecolor=blue,linkcolor=blue,urlcolor=blue,pagebackref]{hyperref}

\usepackage{amsmath}
\usepackage{amssymb}
\usepackage{mathtools}
\usepackage{amsfonts}
\usepackage{bm}
\usepackage{bbm}

\input{Macros/algorithm.tex}
\input{Macros/math.tex}
\input{Macros/theorems.tex}

\input{Macros/misc.tex}

\usepackage[capitalize,noabbrev]{cleveref}

\allowdisplaybreaks

\input{Head/Title}
\input{Head/AuthorInf}

\begin{document}

\maketitle

\input{Main.tex}

\end{document}

%% file: Macros/algorithm.tex
\usepackage{algorithm}
\usepackage{algorithmic}

%% file: Macros/math.tex
\def\1{\bm{1}}

\DeclareMathAlphabet{\mathsfit}{\encodingdefault}{\sfdefault}{m}{sl}
\SetMathAlphabet{\mathsfit}{bold}{\encodingdefault}{\sfdefault}{bx}{n}

\def\calG{{\mathcal{G}}}

\newcommand{\E}{\mathbb{E}}



%% file: Macros/theorems.tex
\usepackage{amsthm}

\theoremstyle{plain}

\newtheorem{theorem}{Theorem}[section]

\newtheorem{proposition}[theorem]{Proposition}

%% file: Macros/misc.tex
\usepackage[textsize=tiny]{todonotes}
\usepackage{multirow}
\usepackage{wrapfig}
\usepackage{subfigure}
\usepackage{tabularx}

\usepackage{xcolor}
\newcount\Comments  
\newcommand{\kibitz}[2]{\ifnum\Comments=1\textcolor{#1}{#2}\fi}

\usepackage{listings}

\usepackage{tabularx}

%% file: Head/Title.tex
\title{Knowledge-Base Gradient Descent: Self-Structuring Learning through Question Answering}

%% file: Head/AuthorInf.tex
\author{Masahiro Kato\thanks{Email: \texttt{mkato-csecon@g.ecc.u-tokyo.ac.jp}}$\,$}

\affil{The University of Tokyo}

\date{\today}

%% file: Main.tex
\begin{abstract}
We consider the problem of constructing knowledge bases for applications based on large language models, in which the structure of a knowledge base determines how an application can retrieve and connect the information needed to answer a question. We study Self-Structuring Learning, a setting in which questions and expected answers guide the construction of a persistent knowledge base whose editable structure can be a graph schema or the organization of linked knowledge documents. We propose Knowledge-Base Gradient Descent, which diagnoses discrepancies between expected answers and responses obtained through the current knowledge base. At each round, the method proposes a structural edit, rebuilds the knowledge base, and adopts the edit when the learning loss decreases. The retained structure is evaluated on other questions about the same sources. By randomizing over candidate edits, we express the loss change of an edit as a derivative with respect to the probability of selecting it, and we evaluate joint edits as single candidates because their changes interact. Schema Descent instantiates the method for graph schemas. In experiments on large documents, each a bundle of research papers, the updated knowledge cards and graph schemas obtain higher mean Answer F1 on held-out generated questions than their initial structures, with schema gains in every document. On human questions, mean F1 rises for schemas and falls for cards: the updated structures serve the kind of question they were learned from.
\end{abstract}


\section{Introduction}

Retrieval-augmented generation (RAG) connects a large language model (LLM) to external knowledge \citep{lewis2020rag}, but the application can fail on a question even when its sources contain the answer. In our experiments, one question asks how much a proposed model improves on the strongest baseline. A knowledge graph built from the paper under an initial schema returns the model's absolute scores and qualitative comparisons, and the application answers that the question is unanswerable. After the schema is changed so that one item keeps a comparison together with its reported improvement and conditions, the application answers with the gains stated in the paper. Whether a question can be answered thus depends on how the knowledge base divides and connects its content.

Constructing a knowledge base requires decisions about the units of knowledge and their connections. A knowledge graph uses a schema to specify entity types and their relationships, and the Open Knowledge Format (OKF) packages knowledge as linked Markdown documents with typed metadata \citep{mcveety2026okf}. Ontology engineering has long used application questions to guide these decisions \citep{gruninger1995methodology}. LLM-based systems now automate parts of construction: GraphRAG builds a graph and community summaries from text, while AutoSchemaKG induces schemas during extraction \citep{edge2025graphrag,autoschemakg2025}. A knowledge base built in either way still has to be judged by the answers it gives, so we take the questions that the application must answer as the objective of construction.

These questions can be prepared before the application is deployed: its specification describes the expected behavior, and during development a model can read the source material directly and help prepare questions with supported answers. When the application answers the same questions through its knowledge base, the differences from the supported answers point to changes in the stored knowledge and its organization. Checking the sources this closely is affordable during development, since the resulting structure and content then serve every later request.

\begin{figure}[t]
\centering
\includegraphics[width=0.9\linewidth]{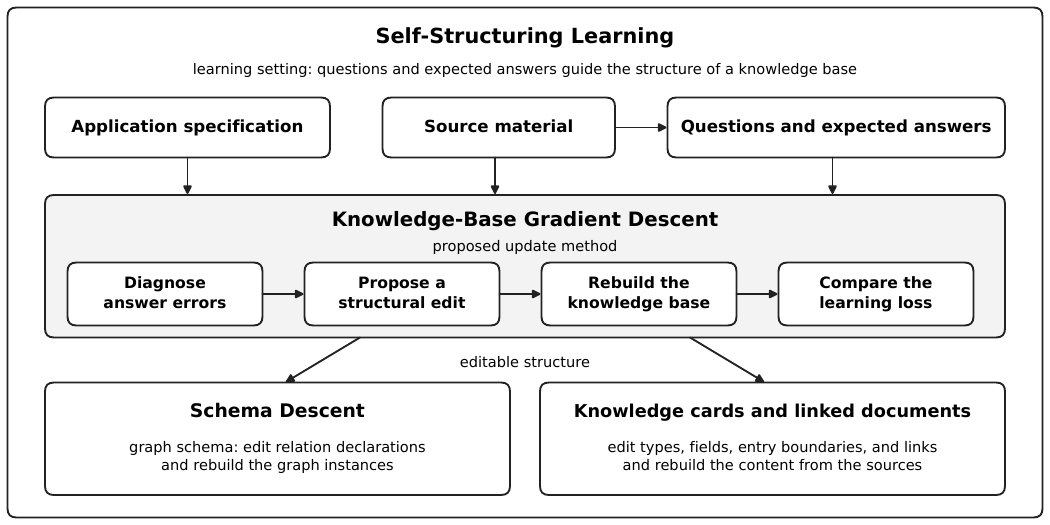}
\caption{Self-Structuring Learning specifies the learning setting. Knowledge-Base Gradient Descent is the proposed update method. Schema Descent applies it to graph schemas, and knowledge cards and linked documents use the same loss comparison with their own structural edits.}
\label{fig:settingmethod}
\end{figure}

We call this learning setting \emph{Self-Structuring Learning}. Within it, we propose \emph{Knowledge-Base Gradient Descent}, which uses answer discrepancies to propose an edit, measures its effect after reconstruction, and adopts it when the learning loss decreases. Repeating this cycle updates the structure used by the application, which can be a graph schema or the organization of linked documents. Figure~\ref{fig:settingmethod} shows how the setting, the method, and its instances relate.

\subsection{Contributions}

Our contribution is a method for learning knowledge base structure from questions and expected answers. Knowledge-Base Gradient Descent separates the diagnosis that proposes an edit from the loss comparison that evaluates it. Each candidate results from one complete structural edit followed by reconstruction, so the comparison measures both repaired answers and errors introduced elsewhere.

We formalize the method through edit directions, which are loss differences between an edited structure and its parent, and relate them to derivatives of a local mixture of candidate losses. Schema Descent is the instance of the method for graph schemas, and its relational model allows structural edits and their interactions to be calculated exactly.

Experiments on five large documents, each a bundle of ten research papers, update the structure of knowledge cards and of a graph schema from questions that the application did not answer as expected. We evaluate the updated structures on held-out generated questions and on human questions about the same papers, and compare them with a baseline that retrieves source passages.

\subsection{Related Work}

GraphRAG and AutoSchemaKG construct a graph or induce its schema from text \citep{edge2025graphrag,autoschemakg2025}, DIAL-KG addresses incremental construction and schema evolution, and TRACE-KG preserves contextual qualifiers and source traceability \citep{bao2026dial,abolhasani2026trace}. These systems obtain the structure from the text and from the knowledge already extracted, whereas we revise it from the answers that an application gives.

Ontology design derives structure from application requirements through competency questions \citep{gruninger1995methodology}. The work closest to ours is \citet{lim2026evaluation}, who treat the schema of generative knowledge graph construction as an optimization target: an evolutionary search varies schemas with textual and structural operators and accepts a candidate when the triples generated under it match reference triples better on validation data. We share their view that a schema can be judged only by what is built under it. Our loss is the answer that an application gives through retrieval, which requires questions and expected answers but no reference triples and applies equally to knowledge cards, and we propose each edit from a diagnosis of the answers that failed. Systems that keep such a structure in use include GBrain, whose schema packs govern extraction, querying, and expert routing, and OKF, which separates the producers of knowledge documents from their consumers \citep{tan2026gbrain,mcveety2026okf}.

Our update rule is related to feedback-guided optimization of prompts and compound systems. ProTeGi generates critiques from a prompt's errors and uses them to propose edits for beam search \citep{pryzant2023}. TextGrad propagates language-model feedback through compound systems \citep{yuksekgonul2025textgrad}. We share with these methods the use of a model's diagnosis of errors as the source of edits, and we apply it to a different object. An edited prompt is used as it is, whereas an edited structure takes effect only after the knowledge base is rebuilt from its sources, so each of our proposals costs a reconstruction. We propose one edit per round and accept it on the whole set of learning questions, where ProTeGi keeps a beam of candidate prompts. ARES uses synthetic data and human annotations to evaluate RAG components \citep{saadfalcon2024ares}; we use reference questions to guide construction as well as to evaluate it.

The graph instance uses established relational representations. Joins describe information recoverable from projections, and view selection allocates resources to useful materialized views \citep{aho1979joins,harinarayan1996cubes}. Relation-instance entities express n-ary associations through binary edges \citep{noy2006nary}, while higher-arity knowledge models preserve richer fact structure \citep{wen2016beyond,rosso2020hinge,galkin2020stare}. Cross-sentence extraction obtains these relations from text \citep{peng2017cross,akimoto2019cross}. Our relational model fixes the source records and edits the declarations that associate their arguments, allowing the effects of those edits to be evaluated directly.

\section{Setup}
\label{sec:setup}

In Self-Structuring Learning, a persistent knowledge structure is learned from questions and expected answers about an application's sources.

\subsection{Application Requirements}

An application specification $\tau$ describes the information needs that the application serves, the requirements for acceptable responses, and any operating constraints. An internal question-answering service, for example, may require answers supported by the organization's documents.

Let $D$ denote a collection of source material and let $x$ denote a question about it. A reference record is $(D,x,y,e)$, where $y$ is an expected answer or a set of scoring requirements and $e$ records the supporting evidence. Evidence may be a passage in $D$ or an explicit behavioral requirement in $\tau$. We write $P_\tau$ for the intended distribution of these records. We prepare questions, expected answers, and supporting evidence according to the specification to approximate this distribution. For learning on one document, $D$ remains fixed while the questions vary.

\subsection{Knowledge Base Structure}

A structure $S$ specifies how source information is organized for repeated question answering. It determines the knowledge units exposed to the application and the relationships and source references retained between them: for a graph, $S$ includes its schema, and for linked documents, it specifies how information is divided into entries and how metadata and links connect those entries.

Let $\mathcal S_\tau$ be the set of structures permitted by the specification. A builder constructs $K_S(D)=\operatorname{Build}(D,S;\xi_B)$, where $\xi_B$ denotes construction randomness. The application produces $\widehat y_S(D,x;\xi)=\operatorname{Answer}(x,K_S(D);\xi_A)$, with answering randomness $\xi_A$ and $\xi=(\xi_B,\xi_A)$. The answering system includes retrieval and may include a language model. The builder and answering procedure remain fixed across structural comparisons, although their constructed knowledge and retrieval indexes change with $S$.

A structural edit must be followed by reconstruction, because a new graph relation requires instances and regrouping knowledge documents requires corresponding content and links. The selected structure remains available for building knowledge from subsequent sources.

\subsection{Reference Questions}

Reference questions may be written by a person, generated by a model reading the sources, or obtained from scenarios in the specification. A factual reference records the answer and its supporting source evidence. For hypothetical scenarios, the stated assumptions provide the reference conditions.

The answering system receives $x$ and access to $K_S(D)$ under the application's rules. The evaluator and diagnostic procedure also receive $y$ and $e$, so the answers obtained through the knowledge base can be compared with the expected answers without supplying them to the answering system.

\subsection{Learning Objective}

We assess a structure through the answers obtained from its constructed knowledge base. Let $\ell(a,y,e)\in[0,1]$ be the application loss for an answer $a$, expected answer $y$, and evidence $e$. The objective is
\begin{equation}
 S^\star\in\operatorname*{arg\,min}_{S\in\mathcal S_\tau}R_\tau(S),\qquad
 R_\tau(S)=\E_{(D,x,y,e)\sim P_\tau,\xi}
 \bigl[\ell(\widehat y_S(D,x;\xi),y,e)\bigr].
 \label{eq:kbobjective}
\end{equation}
The expectation over $\xi$ is taken over the randomness of the fixed builder and answering procedure. The objective in Equation~\eqref{eq:kbobjective} applies to each representation through its own builder and answering procedure.

For a fixed set $Q=\{(D_i,x_i,y_i,e_i)\}_{i=1}^{n}$ of $n$ reference questions and fixed evaluation draws $\xi_i$, define
\begin{equation}
 \widehat R_Q(S)=\frac1n\sum_{i=1}^{n}
 \ell(\widehat y_S(D_i,x_i;\xi_i),y_i,e_i).
 \label{eq:kbsample}
\end{equation}
Questions using the same constructed source may share construction randomness. Common questions and common evaluation draws allow paired comparisons between a parent structure and its edited candidates. When the draws cannot be shared between structures, as in Section~\ref{sec:experiments}, $\xi_i$ in Equation~\eqref{eq:kbsample} denotes the draw realized for the structure being scored.

\section{Knowledge-Base Gradient Descent}
\label{sec:method}

The inputs of Knowledge-Base Gradient Descent are the specification, development sources, an initial structure $S_0\in\mathcal S_\tau$, and the construction and answering procedures of Section~\ref{sec:setup}.

\subsection{Questions for Structural Learning}

We prepare learning questions $Q_f$ to diagnose answer errors and to compare each proposed structure with the current one. Because these questions follow the application's information needs and include questions that the current structure answers correctly, the same set shows whether an edit repairs failures while preserving other answers. Separate questions about the same sources are reserved for assessing the structure retained after learning.

A reference generator, which may read source passages directly and use more computation than the deployed application, records the expected answers and their supporting evidence before structural learning, and the questions and answers then stay fixed during the update cycle.

\subsection{Edits from Answer Discrepancies}

The application answers $Q_f$ through the knowledge base built under the current structure. For each unsuccessful response, we compare it with the expected answer, inspect the retrieved material and the source evidence, and check whether the information needed for the answer was omitted, separated from its context, or linked incorrectly. An LLM can propose these diagnoses using the sources, while the relational model can use source checks and query traces.

Each proposed edit states the affected knowledge units and a change to their organization. For a graph, it may add or replace a relation declaration. For linked documents, it may split or merge entries and update the links that make their context retrievable. When several operations depend on one another, they form one candidate change. Because an answer discrepancy can also reflect an unsupported reference or the answering model's use of retrieved evidence, the effect of a structural edit is measured by comparing the learning loss of the complete reconstructed candidate with that of the current structure.

Formally, a feasible edit $u$ of a parent $S$ produces a structure $u(S)\in\mathcal S_\tau$, and a proposal outside $\mathcal S_\tau$ is rejected. At round $t$, the proposal rule returns one complete edit of the current structure.

\subsection{Measured Edit Directions}

We rebuild the affected knowledge for each proposed structure. For a comparison set $Q$, define the edit direction by
\begin{equation}
 d_Q(S,u)=\widehat R_Q(u(S))-\widehat R_Q(S).
 \label{eq:direction}
\end{equation}
The direction is the loss change of a complete edit, including any required content reconstruction and index updates, and joint changes are evaluated together because their effects need not equal the sum of their individual effects. We adopt the edit when $d_{Q_f}(S,u)<0$.

The method takes its name from a property of the edit direction. When an edit $u$ from a finite set $\mathcal U$ of feasible edits is drawn with probability $\theta_u$ and $S$ is retained otherwise, the expected empirical loss is $\widehat R_Q(S)+\sum_{u\in\mathcal U}\theta_u d_Q(S,u)$, so $d_Q(S,u)$ is its derivative with respect to $\theta_u$ and an edit with a negative direction is a descent direction at $\theta=0$.

\begin{figure}[t]
\centering
\includegraphics[width=\linewidth]{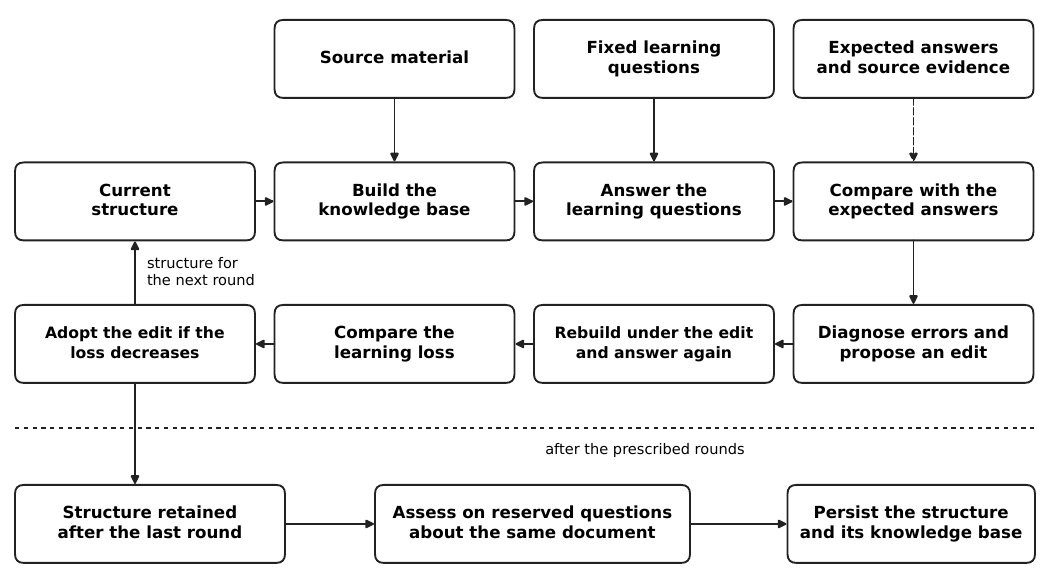}
\caption{The update cycle of Knowledge-Base Gradient Descent and the assessment on reserved questions after the prescribed rounds.}
\label{fig:kbgdflow}
\end{figure}

\subsection{Sequential Candidate Adoption}

Let $S_t$ denote the structure retained after round $t$, with initial structure $S_0$. In each of $T$ rounds, one candidate is proposed from $S_{t-1}$, reconstructed, and scored on $Q_f$; it becomes $S_t$ when its edit direction is negative, and we have $S_t=S_{t-1}$ otherwise. Algorithm~\ref{alg:selfstructuring} and Figure~\ref{fig:kbgdflow} summarize the procedure.

This comparison makes the retained learning loss nonincreasing when the question set and the evaluation draws are fixed. Rejected proposals and their learning scores can inform the next proposal without changing the loss comparison. After $T$ rounds, the output is $S_T$. Reserved questions assess this output against $S_0$ on the same sources.

\begin{figure}[t]
\centering
\includegraphics[width=\linewidth]{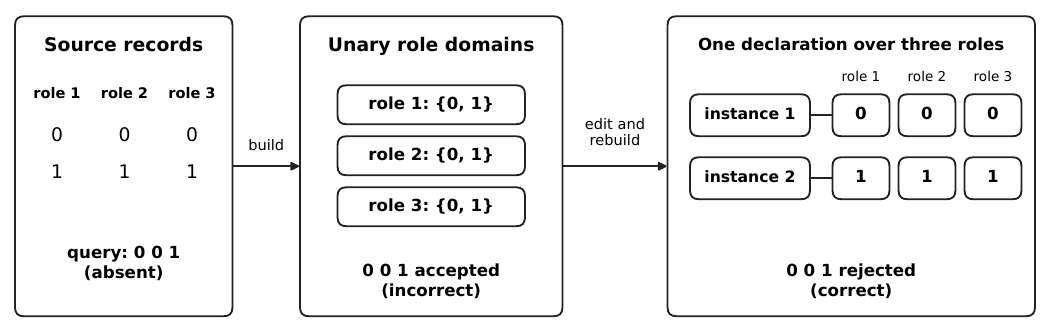}
\caption{A structural edit for the source relation $T_D=\{000,111\}$. Storing only the unary role domains admits the absent tuple $001$. A relation-instance type linking all three roles preserves the two source records and rejects that tuple after reconstruction.}
\label{fig:schemaedit}
\end{figure}

\begin{algorithm}[!htb]
\caption{Knowledge-Base Gradient Descent}
\label{alg:selfstructuring}
\begin{algorithmic}[1]
\REQUIRE specification $\tau$, source collection $D$, learning questions $Q_f$ and reserved questions about $D$, initial structure $S_0\in\mathcal S_\tau$, builder and answering system, edit proposal rule, number of rounds $T$
\STATE Build $K_{S_0}(D)$ and score $S_0$ on $Q_f$
\FOR{$t=1,\ldots,T$}
  \STATE Set $S_t=S_{t-1}$
  \IF{some learning answers have positive loss}
    \STATE Diagnose the unsuccessful answers using source evidence and propose one structural edit $u$ of $S_{t-1}$
    \IF{$u(S_{t-1})\in\mathcal S_\tau$ and the knowledge base rebuilt under $u(S_{t-1})$ is usable}
      \STATE Answer $Q_f$ and compute $d_{Q_f}(S_{t-1},u)$
      \IF{$d_{Q_f}(S_{t-1},u)<0$}
        \STATE Set $S_t=u(S_{t-1})$ and retain its knowledge base and answers
      \ENDIF
    \ENDIF
  \ENDIF
\ENDFOR
\STATE Assess $S_T$ and $S_0$ on reserved questions about the same sources
\STATE Persist $S_T$ and the knowledge base built under it
\RETURN $S_T$ and $K_{S_T}(D)$
\end{algorithmic}
\end{algorithm}

\subsection{Schema Descent}
\label{sec:instance}

Schema Descent applies Knowledge-Base Gradient Descent to graph schemas. In the relational model of Appendix~\ref{app:graph}, the structure $S$ is a set $H$ of relation declarations. A declaration specifies which argument roles share a relation-instance entity. A false association suggests an additional declaration that distinguishes the queried combination from the source records. A tuple that is present in the source records but rejected by incompletely constructed relations suggests declarations to remove or replace. An edit removes a set $U\subseteq H$ and adds a set $V$ of unselected declarations, giving $u(H)=(H\setminus U)\cup V$.

The edit direction of a schema edit is computed by Equation~\eqref{eq:direction} from the edited schema and its reconstructed knowledge. The graph-specific choices are its declaration grammar, construction rule, query procedure, and allowable edits. Figure~\ref{fig:schemaedit} illustrates the change in a stored association. The relational model uses projections to construct the graph and answer membership queries, whereas the experiment in Section~\ref{sec:experiments} edits entity and relation types, extracts entities and relation instances from text under each schema, and answers questions through retrieval.

For knowledge cards or a linked-document knowledge base, the same update rule edits the types, grouping, fields, and links that determine the retrievable knowledge units. An edit preserves source attribution and rebuilds the content governed by the changed structure, and Equation~\eqref{eq:direction} compares the rebuilt candidate with its parent. Section~\ref{sec:experiments} applies the same sequential procedure to knowledge cards and to a graph schema.

\subsection{Interactions between Structural Changes}

For an edit $u$ of $S$, define its empirical gain by $\Delta_Q(S,u)=\widehat R_Q(S)-\widehat R_Q(u(S))=-d_Q(S,u)$. Under exact-answer scoring, write $c_i(S)=\mathbb{I}(\widehat y_S(D_i,x_i;\xi_i)=y_i)$, where $\mathbb{I}$ is the indicator function, so that the loss on record $i$ is $1-c_i(S)$. Then, the gain satisfies
\begin{equation}
 \Delta_Q(S,u)=\frac1n\sum_{i=1}^{n}
 \bigl(\mathbb{I}(c_i(S)=0,\ c_i(u(S))=1)
       -\mathbb{I}(c_i(S)=1,\ c_i(u(S))=0)\bigr).
 \label{eq:kbgain}
\end{equation}
Thus, each repaired answer contributes positively and each newly incorrect answer contributes negatively. General bounded losses use the corresponding paired loss differences.

A joint edit can have a negative direction even when its individual changes do not. If, because their constructed views omit instances, exactly two selected graph declarations reject the same tuple present in the source records, deleting either one leaves the answer incorrect, whereas deleting both corrects it. Knowledge-Base Gradient Descent includes such complete changes as candidates rather than inferring their gain from separate scores. The joint-edit identity in Appendix~\ref{sec:gradients} calculates these interactions for Schema Descent.

\section{Experiments}
\label{sec:experiments}

We apply Knowledge-Base Gradient Descent to a retrieval-based question-answering application over a large document. The experiment asks whether updating the structure of a knowledge base, after some questions are not answered as expected, improves the answers to other questions about the same document. Two representations use the same material: knowledge cards and a graph schema. For each representation, the five documents start from the same initial structure and use the same update rule, retrieval procedure, and models.

\subsection{Documents and Questions}

A large document is a bundle of ten research papers from the Qasper training split, version 0.3 \citep{dasigi2021qasper}, and the experiment uses five such documents. We use each paper's title, abstract, section paragraphs, and figure and table captions from Qasper. A paper is eligible when at least five of its questions have a human answer and its sections or captions contain nonempty text. Eligible papers are put in a fixed order by a hash of the paper identifier (Appendix~\ref{app:retrieval}), and the documents take the papers of that order ten at a time, without using answer scores for selection.

Each document is built into one knowledge base with one retrieval index, and one structure applies to the whole document. The five documents share no paper and are updated independently. Since Qasper questions presuppose a single paper, each question given to the answerer names its paper by title.

Evaluation uses every human question about the bundled papers, with all of its annotations; the five documents have 62, 58, 54, 53, and 66 such questions, 293 in total. These questions are used only for evaluation.

The learning questions and the generated evaluation questions are written by a model. Following the way Qasper questions were collected, a question writer sees only the title and abstract of a paper and writes five questions about its content. A separate reference writer reads the whole text of the paper and writes the expected answer to each question: the shortest sufficient answer, Yes or No for a yes-or-no question, or Unanswerable when the paper does not state the answer. For an answerable question, the reference writer also gives a short quotation from the paper that supports the answer. For the paper at zero-based position $k$ in a document, the question at zero-based position $k\bmod5$ is held out as a generated evaluation question, and the other four join the learning set $Q_f$. Each of the five question positions is therefore held out twice per document, which has 40 learning questions and 10 generated evaluation questions. Agreement with the generated expected answers and agreement with the human annotations are reported separately.

\subsection{Structures and Construction}

The structure $S$ specifies the units into which the document is divided. In the card representation, the structure is a list of card types. A type has a name, a definition of a card's content, a grouping rule that fixes the scope of one card, and at most five named fields; the definition and the grouping each have at most 500 characters. The initial card structure has one type, with one card per section that summarizes the section.

In the schema representation, the structure is a graph schema: a list of entity types and a list of relation types, where every type has a definition of at most 500 characters and a relation type names two to five argument roles. The initial schema has five entity types (method, dataset, metric, task, and setting) and five relation types (a method uses a dataset; a method is evaluated with a metric on a dataset, with the reported value; two methods are compared, with the outcome; a method addresses a task; a method is trained or run with a setting, with its value). In both representations, a structure has one to ten types of each kind; Appendix~\ref{app:retrieval} gives the constraints on names.

The builder receives only the structure and one paper, and writes cards under the card types, or entities and relation instances under the entity and relation types. The text of a card or of a relation instance may quote or summarize the paper. We call each card, entity, or relation instance an item.

A construction reply that cannot be parsed or that violates the structure is unusable as a whole; Appendix~\ref{app:retrieval} lists the conditions. An unusable initial construction stops the experiment; an unusable candidate reconstruction is rejected without a learning score and preserves the current structure.

The retrieved evidence names the paper once, in a line before the items. The knowledge base consists only of the items produced under the structure, and it is built again from the papers when the structure changes.

\subsection{Retrieval and Answering}
\label{sec:retrieval}

All items of a document are indexed together. For a question, BM25 ranks the items of its paper by the words of the question, and at most 50 candidates are kept. The query is the question without the title of its paper, after a fixed list of stop words is removed; Appendix~\ref{app:retrieval} gives the list and the order of items with equal scores. The line naming the paper is counted once, and items are then passed to the answerer in rank order within 1,000 units in total, where a unit is a word, a number, or any other non-space character. In the schema representation, a candidate relation instance is followed by the entities it links, and a candidate entity by its relation instances, within the same budget.

The answerer receives only the question, the line naming the paper, and the retrieved items. It returns the shortest sufficient answer, Yes or No, or Unanswerable. For each answer, we take the largest official Qasper Answer F1 over its reference answers and define the loss as one minus that value.

\subsection{Updates}

Five update rounds are run for each representation of each document, and each round answers $Q_f$ through the current knowledge base. A learning question fails when its F1 is below one. The proposer receives the current structure with statistics on the use of its types, up to ten failed questions with their expected and actual answers and the items passed to the answerer, and every earlier rejected proposal with its learning F1 and the reason for rejection; Appendix~\ref{app:retrieval} gives the full input. It returns one complete new structure, which defines the edit $u$ of Section~\ref{sec:method}.

A reply that violates the constraints on a structure is returned once with every violation listed. If the second reply still violates them, or a reply changes the kind of structure or repeats the current structure, the current structure stays in place.

The knowledge base is built again under the proposed structure, and $Q_f$ is answered through it. We adopt the proposal when the mean learning F1 increases, that is, when $d_{Q_f}(S,u)<0$, as in Algorithm~\ref{alg:selfstructuring}. The experiment evaluates the structure retained after the last round on the human and generated evaluation questions of the same document.

Question writing, reference writing, construction, and proposals use the model alias \texttt{gpt-6.1-sol}; answering uses \texttt{gpt-6-luna}. Reasoning effort is medium for question writing, reference writing, and proposals, and low for construction and answering. Each distinct request has one saved response, which is reused whenever the same request recurs, so the edit direction of a proposal compares the saved responses of the two structures on the same questions.

\subsection{Comparisons}

The primary comparison is the Answer F1 of the updated structure minus that of the initial structure on the human evaluation questions, reported for cards and for the schema; the same difference on the generated evaluation questions is reported beside it. For a document, Answer F1 is averaged first over the questions of each paper and then equally over the papers, and we report the difference of every document together with the equal mean of the five document differences. For each document, we give a 95\% percentile bootstrap interval from 10,000 paired resamples of its papers, which describes variation across papers.

As a baseline outside the two compared representations, we split each paragraph of the source text into passages of at most 100 units, prefix each passage with its section name, and retrieve them under the same procedure and 1,000-unit budget. The baseline is scored on the same human and generated evaluation questions.

\subsection{Results and Discussion}
\label{sec:empirical-results}

Table~\ref{tab:structure-results} reports the means that weight the five documents equally. On the human questions, the card updates lower mean F1 by 2.89 points and the schema updates raise it by 1.10 points; on the generated evaluation questions, the two gains are 4.45 and 12.91 points. Schema updates therefore produce the larger mean improvement on both evaluation sets, while the updated cards have the higher absolute F1 on both.

\begin{table}[t]
\centering
\caption{Answer F1 in points, averaged equally over the five documents. Differences are updated minus initial.}
\label{tab:structure-results}
\begin{tabular}{lrrrrrr}
\toprule
 & \multicolumn{3}{c}{Human} & \multicolumn{3}{c}{Generated} \\
\cmidrule(lr){2-4}\cmidrule(lr){5-7}
Representation & Initial & Updated & Difference & Initial & Updated & Difference \\
\midrule
Cards & 35.56 & 32.67 & $-2.89$ & 50.61 & 55.06 & $+4.45$ \\
Schema & 28.97 & 30.07 & $+1.10$ & 35.33 & 48.24 & $+12.91$ \\
\bottomrule
\end{tabular}
\end{table}

\begin{figure}[t]
\centering
\includegraphics[width=\linewidth]{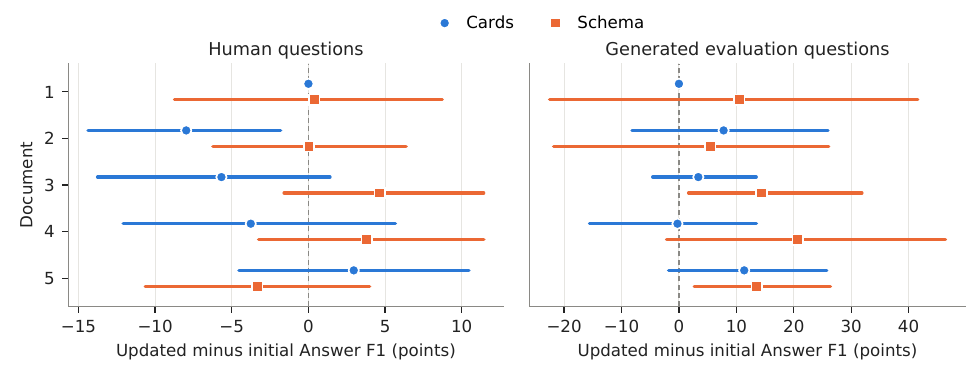}
\caption{Document-level Answer F1 differences, updated minus initial, in points. Each mark is the difference of one document, and each bar is its 95\% bootstrap interval over papers. The dashed line marks no change, and the two panels have different horizontal scales.}
\label{fig:documents}
\end{figure}

Figure~\ref{fig:documents} shows the document-level differences with their bootstrap intervals, and Table~\ref{tab:structure-documents} in Appendix~\ref{app:results} lists the values. On the generated questions, schemas improve in every document, by 5.46 to 20.66 points, and cards in three of the five. On the human questions, schemas improve in four documents and cards in one. The intervals exclude zero for the schema gains on generated questions in documents 3 and 5 and for the card decline on human questions in document 2.

The human-question and generated-question differences have opposite signs for cards in documents 2 and 3 and in the mean, and for the schema in document 5. The two evaluation conditions differ in both the questions asked and the source of their reference answers: generated questions use the reference writer's answers, whereas human questions use the Qasper annotations.

\begin{figure}[t]
\centering
\includegraphics[width=\linewidth]{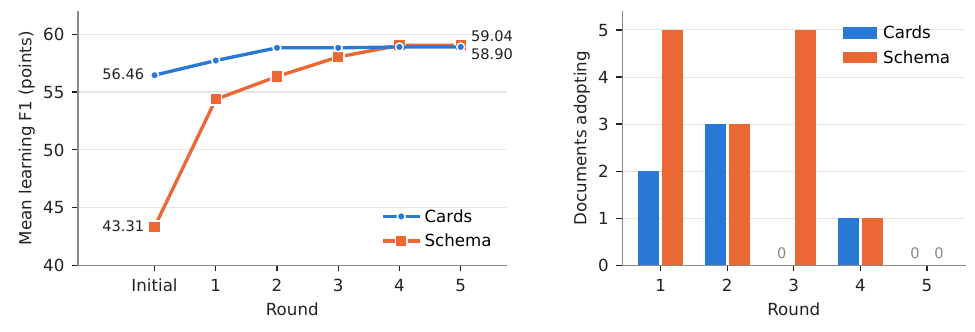}
\caption{Left: mean learning F1 in points of the structure retained after each round, averaged equally over the five documents. Right: the number of documents that adopt the proposal of each round.}
\label{fig:rounds}
\end{figure}

Figure~\ref{fig:rounds} follows the retained structures through the five update rounds, and Table~\ref{tab:structure-learning} in Appendix~\ref{app:results} lists the values. Mean learning F1 increases from 56.46 to 58.90 for cards and from 43.31 to 59.04 for schemas. Of the 25 proposals for each representation, one per round in each of the five documents, six card proposals and fourteen schema proposals are adopted. The numbers of adopted proposals in documents 1 through 5 are 0, 2, 2, 1, and 1 for cards, and 3, 4, 3, 2, and 2 for schemas. One card proposal and four schema proposals are rejected before learning scores are obtained: two proposals still violate the constraints after repair, and three reconstructions are unusable. No proposal is adopted in the fifth round.

The source-passage baseline scores 37.03 on human questions and 46.71 on generated evaluation questions. Its human score exceeds the scores of both updated representations, while its generated score is below both.

The learned structures change what a knowledge unit covers and how its information is organized. In document 2, the initial card structure has one section-summary type with no named fields; the retained structure has eight types, including dataset, protocol, and result cards. A protocol card keeps a complete evaluation or controlled comparison together across source sections, including the training, development, test, and held-out roles. Its five fields record the task and datasets; the procedure and data roles; controls, definitions, and restrictions; the scope of the evidence; and the source and links. A result card keeps a comparison with its datasets, metrics, baselines, and exceptions. The same structure governs all ten papers, and rebuilding the knowledge base under it produces 249 cards, compared with 210 under the initial structure.

In document 3, the schema changes from five entity types and five relation types to ten entity types and three relation types. The new \texttt{finding} type keeps the scores of the winning model, reported improvement, metric, baseline, and experimental conditions together in one comparison. The \texttt{result\_detail} relation links a method or task to a finding, while \texttt{evidence\_context} links an evaluation protocol to its finding; both have the roles \texttt{subject} and \texttt{evidence}. The total number of entities and relation instances changes from 937 to 563.

The saved answers also show changes in the evidence delivered to the answerer. In the schema representation of document 3, a generated question asks how much Stacked DeBERT improves on the strongest baseline for each dataset. The initial retrieved items give absolute scores and qualitative comparisons, but omit the requested comparison figures. The updated retrieved items give the reported 6\% to 8\% gains for Twitter and the 0.94\% and 1.89\% gains for the two Chatbot conditions. The answer changes from Unanswerable to these figures, and Answer F1 rises from 0 to 80. In the card representation of document 2, a human question asks for the definition of hate speech. The cards passed to the answerer under the initial structure include the explicit definition, while those passed under the updated structure omit it; the answer changes to Unanswerable and F1 falls from 82.61 to 0.

Answer F1 also responds to wording. For a human question about data augmentation answered through the schema of document 5, the initial answer is ``Back translation between English and Chinese.'' and the updated answer is ``English--Chinese back-translation''. The Qasper reference is ``back translation between English and Chinese''. The answers express the same augmentation technique, and the updated retrieved evidence still states it, but their Answer F1 values are 100 and 0. The evaluator keeps the en dash in ``English--Chinese'' and removes the hyphen in ``back-translation'', producing the tokens ``english--chinese'' and ``backtranslation''. Neither token matches a separate word in the reference. This question contributes $-1.67$ points to the schema's document-level human difference of $-3.34$.

The update fits the structure to the learning questions. A model writes these questions from a title and an abstract, the generated evaluation questions are written in the same way, and both representations improve on them. The human questions come from a different source, and on them an update redistributes the answers: of the 293 human questions, the card updates raise the F1 of 53 and lower that of 66, and the schema updates raise the F1 of 80 and lower that of 73.

The schema updates make the items fewer and larger, so that an item keeps a finding together with its conditions, as in document 3: the five documents hold 4,384 items under the initial schemas and 2,756 under the updated ones, and the answerer receives 18.1 items per human question instead of 25.6 within the same budget. With these items the answerer answers questions that it had declined, and the number of Unanswerable answers falls from 131 to 95 on the human questions and from 32 to 16 on the 50 generated evaluation questions.

The card updates divide the content further: the initial cards summarize each section, and the updates distribute this content among typed cards, from 877 cards to 1,328 over the five documents, so that fewer cards fit in the evidence for a question (7.7 instead of 8.4 on the human questions). A typed card serves the questions that its type anticipates, as the protocol and result cards of document 2 do, and a question outside these types can lose the sentence it needs, as the definition of hate speech shows.

Source passages keep the wording of the paper and leave nothing out, which serves the human questions, while the learned structures do better on questions written like the learning questions, where both exceed the baseline. The baseline also declines to answer more reliably: on the 37 human questions whose annotations are all Unanswerable, it scores 86.5, against 78.4 for both initial representations and 70.3 for both updated ones. We conclude that the learning questions determine what a learned structure serves, and that they should be drawn from the questions the application will receive, as the objective in Equation~\eqref{eq:kbobjective} prescribes.

\section{Conclusion}

Self-Structuring Learning uses an application's expected behavior to define how a persistent knowledge base should be organized. Knowledge-Base Gradient Descent updates the structure from the answers that the application gets wrong and keeps an edit only when the rebuilt knowledge base lowers the learning loss. Each edit direction is the loss change of a complete edit, including interactions between structural operations, and Schema Descent realizes the method for graph schemas, with a relational model in which these effects can be calculated exactly. In the retrieval experiment on five documents, updates improve mean F1 on held-out generated questions for both knowledge cards and graph schemas, with schema improvements in every document, while on human questions mean F1 increases for schemas and decreases for cards. The updated structures serve the kind of question they were learned from, so the learning questions should represent the questions of the application.

\bibliography{arXiv2.bbl}

\bibliographystyle{tmlr}

\onecolumn

\appendix

\section{Graph Instantiation}

\label{sec:model}
\label{app:graph}

A relational model gives a mathematical instance of Schema Descent. Projections define its constructed graph and membership answers, so the effects of schema edits have exact expressions.l

\subsection{Relational Model}

For each source collection $D$, let $T_D$ be the corresponding finite table; all tables have the same $d$ known argument roles. Write $\pi_A$ for projection onto a role set $A$ and define $\mathcal V_{D,j}=\pi_{\{j\}}(T_D)$ for $j\in\{1,\ldots,d\}$. Thus, we have $T_D\subseteq\prod_{j=1}^{d}\mathcal V_{D,j}$. Each row is a complete source record for the membership task, so absence from $T_D$ gives a negative reference under the declared closed-world semantics.

Let $\calG=\{A\subseteq\{1,\ldots,d\}:|A|\geq2\}$ be the declaration grammar. A schema $H\subseteq\calG$ specifies a relation-instance type for each $A\in H$, together with one typed argument edge for each role in $A$. Under exact construction, every distinct row in $\pi_A(T_D)$ becomes an instance of that type. Identifiers are local to the source and declaration, so they cannot link different views through a shared source-record identifier. Before any relation declaration is added, the initial schema stores only the unary role domains.

A membership query is $x=(D,z)$, where $z=(z_1,\ldots,z_d)\in\prod_j\mathcal V_{D,j}$. Its reference is $y=\mathbb{I}(z\in T_D)$. We call the query and its tuple source-positive when $y=1$ and source-negative when $y=0$. The query procedure accepts exactly the tuples in
\begin{equation}
 J_H(T_D)=\{z\in\textstyle\prod_j \mathcal V_{D,j}:\pi_A(z)\in\pi_A(T_D)\text{ for every }A\in H\}.
 \label{eq:join}
\end{equation}
The empty schema accepts the Cartesian product of the domains. Including the full role set $\{1,\ldots,d\}$ as one declaration reproduces $T_D$ exactly. Intermediate schemas can admit spurious joins. If construction omits relation instances, we replace $\pi_A(T_D)$ in Equation~\eqref{eq:join} with its constructed subset $\widetilde T_{D,A}$, which we call the constructed view of $A$; under exact construction, the view is $\pi_A(T_D)$ itself. For each comparison, we fix the constructed views and reuse the same view for a given source and declaration in every candidate schema.

The learner in this model receives source tables and membership queries with reference answers. For each declaration that the learner requests, it observes whether the constructed view rejects each query, and it uses these responses for diagnosis and loss comparison. The sequential procedure compares each candidate with its parent on these queries.

The declaration set $H$ instantiates $S$ in Section~\ref{sec:setup}. Choosing $H$ is a view-selection problem in which the selected projections may admit tuples absent from $T_D$.

\subsection{Query Cost}
\label{sec:cost}

For the exact view $\pi_A(T_D)$ of a declaration $A$, let $I_{D,A,j}(v)$ be the posting list of its relation-instance identifiers whose argument at role $j$ equals $v$. Membership in that view is determined by intersecting the lists $I_{D,A,j}(z_j)$ over $j\in A$. A fixed eager plan fetches every required list and evaluates every selected view. The number of adjacency lookups plus identifiers fetched is
\begin{equation}
 w_A(D,z)=|A|+\sum_{j\in A}|I_{D,A,j}(z_j)|,
 \qquad w_H(D,z)=\sum_{A\in H}w_A(D,z).
 \label{eq:work}
\end{equation}
The fixed unary-domain checks are common to all schemas and are not counted. This definition measures query-time accesses in the graph plan just defined, and query work is additive across views because that plan performs no early termination or cross-view sharing.

For this cost model, an independent set $C$ of unlabeled calibration queries gives $c_C(A)=|C|^{-1}\sum_{(D,z)\in C}w_A(D,z)$ and $c_C(H)=\sum_{A\in H}c_C(A)$. For a fixed source, if each role of $A$ has $q$ values and the queried value of each role is uniform over them, the mean of $w_A$ is $|A|+|A||\pi_A(T_D)|/q$. For fixed $q$, this mean increases with $|A|$ when $|\pi_A(T_D)|$ is fixed, and with $|\pi_A(T_D)|$ when $|A|$ is fixed. Since the work is defined on the exact view, omissions in a constructed view do not make a schema appear cheaper.

Let $P$ be a distribution of membership queries and their reference answers. Write $a_H(D,z;\xi_B)$ for the graph answer with construction randomness $\xi_B$. For a budget $B\geq0$, we aim to select
\begin{equation}
 H^\star\in\operatorname*{arg\,min}_{H\subseteq\calG:\,c_C(H)\leq B}
 R_P(H),\qquad
 R_P(H)=\E_{(D,z,y)\sim P,\xi_B}[\mathbb{I}(a_H(D,z;\xi_B)\neq y)].
 \label{eq:objective}
\end{equation}
The budget constrains calibrated mean work under the eager plan.

\subsection{Schema Descent Updates}

\label{app:schemadescent}

Schema Descent diagnoses the current graph, proposes feasible changes to its declarations, and compares the edited schemas with their parents.

\subsubsection{Diagnosing Incorrect Associations}

For a source-negative tuple $z$, which is a false association of the schema $H$ when $z\in J_H(T_D)$, define
\begin{equation}
 \mathcal W_D(z)=\{A\in\calG:\pi_A(z)\notin\pi_A(T_D)\}.
 \label{eq:repairs}
\end{equation}
Each member is a declaration that rejects $z$ under exact construction; we say that it separates $z$ from the source records. A single false association can therefore be repaired by several different declarations.

For each $r\in T_D$, let $\mathcal M_r(z)=\{j:z_j\neq r_j\}$ be its mismatch set. The following characterization obtains the separating declarations from the source table alone, without the constructed views, and connects them to hitting-set enumeration in data profiling \citep{blasius2022hitting}.
\begin{proposition}[Separating declarations]
\label{prop:separators}
A declaration $A\in\calG$ belongs to $\mathcal W_D(z)$ if and only if it intersects every set in $\{\mathcal M_r(z):r\in T_D\}$. The family $\mathcal W_D(z)$ is upward closed. Every hitting set of these mismatch sets has at least two roles, so the inclusion-minimal members of $\mathcal W_D(z)$ are the minimal hitting sets.
\end{proposition}
\begin{proof}
The projection of $z$ onto $A$ is absent exactly when every source record differs from $z$ on some role in $A$. This is the hitting-set condition. If this condition holds for $A$, it also holds for every superset of $A$. For each role $j$, the value $z_j$ lies in $\mathcal V_{D,j}$, so some source record $r$ satisfies $r_j=z_j$ and $j\notin\mathcal M_r(z)$; hence no single role is a hitting set.
\end{proof}

\subsubsection{Joint Structural Repairs}

A minimal separator of each failure can be worse than one larger declaration that repairs several failures at once, even under the query-work costs of Appendix~\ref{sec:cost}.
\begin{proposition}[A nonminimal joint repair]
\label{prop:larger}
There are source relations and query-work budgets for which using only minimal separators of the observed failures gives strictly greater minimum empirical error than allowing their supersets.
\end{proposition}
\begin{proof}
Consider one source $D$ with $T_D=\{000,111\}$ and the sample of the three source-negative membership queries $001$, $010$, and $100$. Each pair declaration rejects two of these queries, and the full triple rejects all three. Each stored role list has length one for any query, giving work $4$ for a pair and $6$ for the triple under every calibration set. A budget of $6$ admits the triple but no schema with two pairs. Proposals restricted to minimal separators incur error $1/3$, whereas the full triple attains zero error. The triple is nonminimal for each individual failure.
\end{proof}

\subsubsection{Signed Edit Evaluation}

A schema edit can add a declaration or remove a declaration whose constructed view rejects source-positive tuples. For a labeled membership query $i$, let $b_{i,A}$ indicate rejection by the constructed view of $A$, and define the blocking set $\mathcal B_i(H)=\{A\in H:b_{i,A}=1\}$. The schema accepts exactly when $\mathcal B_i(H)=\varnothing$. Assign $s_i=1-2y_i$, so a newly rejected source-negative query improves the answer while a newly rejected source-positive query worsens it.

We write a schema edit as $(U,V)$, where $U\subseteq H$ and $V\subseteq\calG\setminus H$. Its result is $H'=(H\setminus U)\cup V$, subject to $c_C(H')\leq B$. For a fixed sample of $n$ membership queries, write $a_i(H)=\mathbb{I}(\mathcal B_i(H)=\varnothing)$ for the answer on query $i$ and $\widehat R(H)=n^{-1}\sum_{i=1}^n\mathbb{I}(a_i(H)\neq y_i)$ for the empirical risk. The edit gain is $\Delta_H(U,V)=\widehat R(H)-\widehat R(H')$. Comparing edits on the same sample accounts for both repaired answers and answers damaged by a change.

Under exact construction, every source-positive tuple is accepted. A rejected source-positive query therefore shows that some constructed view omits instances, and the declarations in its blocking set are the candidates for removal; a false association gives no such information. Removing a declaration can also admit a source-negative tuple that no remaining declaration rejects. If a source-positive tuple is rejected by several selected declarations, deleting only one may leave its answer unchanged. Removing $\mathcal B_i(H)$ releases it unless an added declaration rejects it again. Appendix~\ref{sec:gradients} relates single edits to exact coordinate gradients and characterizes the gain of a joint edit.

\subsection{Coverage under Exact Construction}

Every source-positive tuple passes every exact projection. Let $N_A$ be the event that a membership query is source-negative and rejected by the exact view of $A$. Then, the reduction in error from adding declarations is
\begin{equation}
 F(H)=\mathbb{P}\Bigl(\bigcup_{A\in H}N_A\Bigr),\qquad
 R_P(H)=\mathbb{P}(y=0)-F(H).
 \label{eq:coverage}
\end{equation}
The union measure is monotone and submodular, and its finite-sample optimization under the budget is budgeted maximum coverage \citep{khuller1999coverage}.

\begin{proposition}[Empirical sufficiency of all sample repairs]
\label{prop:complete}
Under exact construction, collect $\mathcal W_D(z)$ for every source-negative query in a fixed sample. With positive additive declaration costs, the best feasible empirical error over the union of these sets equals the best feasible empirical error over the full grammar.
\end{proposition}
\begin{proof}
A declaration outside that union rejects none of the source-negative sampled queries. Exact construction also prevents it from rejecting a source-positive query. Removing such a declaration preserves every sampled prediction and reduces cost. Applying this removal to a full-grammar optimum gives a feasible optimum in that union.
\end{proof}

\subsection{Gradients for Schema Edits}
\label{sec:gradients}

For a fixed pool $\mathcal P\subseteq\calG$ of $m$ declarations, consider inclusion probabilities $p\in[0,1]^m$. Independent inclusion gives the empirical expected loss
\begin{equation}
 \mathcal L(p)=\frac1n\sum_{i=1}^n\left(y_i+s_i\prod_{A\in\mathcal P}(1-p_A b_{i,A})\right).
 \label{eq:multilinear}
\end{equation}
This is the standard multilinear extension of the empirical set function \citep{chekuri2014multilinear}. At the indicator vector $\mathbf1_H$ of a schema $H\subseteq\mathcal P$, it equals $\widehat R(H)$.

The function $\mathcal L$ is affine in each $p_A$, so for each $A\in\mathcal P$ its coordinate gradient at $\mathbf1_H$ is
\begin{equation}
 \begin{aligned}
 g_A(H)&=\partial_A\mathcal L(\mathbf1_H)
 =-\frac1n\sum_{i=1}^n s_i b_{i,A}
      \prod_{A'\in\mathcal P\setminus\{A\}}(1-\mathbf1_H(A')b_{i,A'})\\
 &=\widehat R(H\cup\{A\})-\widehat R(H\setminus\{A\}).
 \end{aligned}
 \label{eq:gradient}
\end{equation}
The risk reduction is $-g_A(H)$ for adding an unselected declaration and $g_A(H)$ for deleting a selected declaration. For a feasible single addition or deletion, the edit direction $d_Q$ in Equation~\eqref{eq:direction} therefore equals $g_A(H)$ for the addition and $-g_A(H)$ for the deletion on the same comparison sample.

The derivative explains the direction of a single change, whereas replacements and other joint updates must account for interactions among views. For an addition set $V$, let $b_i(V)=\mathbb{I}(\exists A\in V:b_{i,A}=1)$.

\begin{proposition}[Exact gain of a joint edit]
\label{prop:editgain}
For any feasible edit $(U,V)$ of $H$, the empirical risk reduction is
\begin{equation}
 \Delta_H(U,V)
 =\frac1n\sum_{i:\mathcal B_i(H)=\varnothing}s_i b_i(V)
 -\frac1n\sum_{i:\varnothing\neq \mathcal B_i(H)\subseteq U}s_i(1-b_i(V)).
 \label{eq:editgain}
\end{equation}
The identity holds for arbitrary fixed constructed-view responses, including omissions.
\end{proposition}
\begin{proof}
An accepted query becomes rejected exactly when an added view rejects it, contributing the first sum. A rejected query becomes accepted exactly when every previously blocking declaration is removed and no added declaration rejects it, contributing the second sum with opposite sign. Every other query keeps its previous answer, so summing these signed changes gives the risk reduction.
\end{proof}

This identity covers corrections that a coordinate gradient cannot express alone. If exactly two selected views reject a source-positive tuple, its contribution to either single-deletion gradient is zero because the other view still rejects it. Deleting both removes that false negative. Similarly, the example in Appendix~\ref{app:noise} has two harmful individual additions whose joint addition reduces risk. Schema Descent evaluates these compound edits directly, using the loss difference between the complete edited schema and its parent, as in Knowledge-Base Gradient Descent.

\subsection{Construction Errors}
\label{app:noise}

Let $\widetilde N_A$ be the event that a constructed view rejects a source-negative query, and let $\widetilde P_A$ be the event that it rejects a source-positive query. The exact improvement of the realized graph over the unary representation is
\begin{equation}
 \widetilde F(H)=\mathbb{P}\Bigl(\bigcup_{A\in H}\widetilde N_A\Bigr)
 -\mathbb{P}\Bigl(\bigcup_{A\in H}\widetilde P_A\Bigr).
 \label{eq:signed}
\end{equation}
Thus, omissions lead to a difference of coverage functions. A declaration may eliminate incorrect combinations while rejecting some source-positive tuples, and the objective can lose both monotonicity and submodularity.

For example, consider one source $D$ with $T_D=\{0000,0101,1010\}$ and the declarations $A=\{1,2\}$ and $A'=\{3,4\}$. Both exact projections are $\{00,01,10\}$. Omit $00$ from both constructed views and retain the other instances. Give probability $0.4$ to the source-positive query $0000$, and probability $0.3$ each to the source-negative queries $1101$ and $0111$. Both constructed views reject the source-positive query, while only that of $A$ rejects $1101$ and only that of $A'$ rejects $0111$. Then, we have $\widetilde F(\{A\})=\widetilde F(\{A'\})=-0.1$ and $\widetilde F(\{A,A'\})=0.2$. The marginal gain from adding $A'$ is $-0.1$ at the empty schema and $0.3$ when $A$ is already selected, so $\widetilde F$ is neither monotone nor submodular.

A declaration can therefore increase loss when its constructed view omits instances, and loss comparison uses the constructed graph responses for this reason.

\section{Experimental Details}
\label{app:retrieval}

Eligible papers are ordered by the SHA-256 value of the UTF-8 string \texttt{0|} followed by the paper identifier. Every type, field, and role name has at most 50 characters, begins with a lowercase letter, consists of lowercase letters, digits, and underscores, and is unique within its list.

A construction reply is unusable as a whole when it cannot be parsed, names a type outside the structure or a section outside the paper, or contains a card with neither text nor a filled field, an entity without a name or an identifier, two entities with the same identifier, or a relation with an empty role. A duplicate identifier is excluded because a relation could otherwise refer to more than one entity. A reply cut off at the model's output limit is also unusable, as is a knowledge base in which the ten papers yield no item.

The query of Section~\ref{sec:retrieval} consists of the words of the question, in lowercase, that are not among the following stop words: a, an, the, of, in, on, at, to, for, from, by, with, and, or, but, is, are, was, were, be, been, being, do, does, did, what, which, who, whom, whose, when, where, why, how, this, that, these, those, it, its, they, their, them, paper, as, not, no, into, than, then, there, here, also, can, could, would, should, have, has, and had. BM25 scores are rounded to ten decimal places, and items with equal scores are ordered by identifier. If fewer than 50 items share a word with the query, items sharing none follow in identifier order. A candidate that does not fit in the budget is skipped together with its links, and later candidates are still considered. In the schema representation, links are followed one step, a linked item that does not fit is skipped, and no item is passed twice.

The proposer receives the current structure and, for each type with at least one item, the number and mean length of its items and the number of times they were passed to the answerer over the learning questions, together with the number of all items, their total length, and the mean number of items passed per learning question. It also receives up to ten failed questions, those with the lowest F1 first and ties in identifier order. Each failed question comes with its expected and actual answers, the items passed to the answerer, a source quotation when available, and, for an answerable reference, the first BM25 match to that quotation and expected answer when found, together with whether that item reached the answerer. Every earlier rejected proposal comes with its learning F1 when available and the reason for rejection.

The bootstrap resamples the ten papers of a document in pairs and uses seed zero.

\section{Detailed Results}
\label{app:results}

Tables~\ref{tab:structure-documents} and~\ref{tab:structure-learning} list the values shown in Figures~\ref{fig:documents} and~\ref{fig:rounds}.

\begin{table}[!ht]
\centering
\caption{Document-level Answer F1 differences in points and their 95\% bootstrap intervals over papers.}
\label{tab:structure-documents}
\begin{tabular}{llrlrl}
\toprule
 & & \multicolumn{2}{c}{Human} & \multicolumn{2}{c}{Generated} \\
\cmidrule(lr){3-4}\cmidrule(lr){5-6}
Document & Representation & Difference & 95\% interval & Difference & 95\% interval \\
\midrule
\multirow{2}{*}{1} & Cards & $0.00$ & $[0.00, 0.00]$ & $0.00$ & $[0.00, 0.00]$ \\
 & Schema & $+0.38$ & $[-8.74, 8.73]$ & $+10.53$ & $[-22.53, 41.60]$ \\
\addlinespace
\multirow{2}{*}{2} & Cards & $-7.97$ & $[-14.37, -1.81]$ & $+7.78$ & $[-8.18, 25.96]$ \\
 & Schema & $+0.03$ & $[-6.23, 6.37]$ & $+5.46$ & $[-21.83, 26.08]$ \\
\addlinespace
\multirow{2}{*}{3} & Cards & $-5.67$ & $[-13.75, 1.40]$ & $+3.38$ & $[-4.57, 13.45]$ \\
 & Schema & $+4.64$ & $[-1.58, 11.43]$ & $+14.37$ & $[1.69, 31.87]$ \\
\addlinespace
\multirow{2}{*}{4} & Cards & $-3.76$ & $[-12.09, 5.67]$ & $-0.26$ & $[-15.59, 13.49]$ \\
 & Schema & $+3.81$ & $[-3.25, 11.44]$ & $+20.66$ & $[-2.16, 46.42]$ \\
\addlinespace
\multirow{2}{*}{5} & Cards & $+2.96$ & $[-4.54, 10.47]$ & $+11.37$ & $[-1.80, 25.75]$ \\
 & Schema & $-3.34$ & $[-10.65, 3.99]$ & $+13.53$ & $[2.66, 26.38]$ \\
\bottomrule
\end{tabular}
\end{table}

\begin{table}[!ht]
\centering
\caption{Mean learning F1 in points of the structure retained after each round, and the number of documents adopting that round's proposal.}
\label{tab:structure-learning}
\begin{tabular}{llrrrrrr}
\toprule
Representation & Quantity & Initial & Round 1 & Round 2 & Round 3 & Round 4 & Round 5 \\
\midrule
\multirow{2}{*}{Cards} & F1 & 56.46 & 57.73 & 58.82 & 58.82 & 58.90 & 58.90 \\
 & Adopted & -- & 2 & 3 & 0 & 1 & 0 \\
\addlinespace
\multirow{2}{*}{Schema} & F1 & 43.31 & 54.37 & 56.34 & 58.04 & 59.04 & 59.04 \\
 & Adopted & -- & 5 & 3 & 5 & 1 & 0 \\
\bottomrule
\end{tabular}
\end{table}